%% file: main.tex
\documentclass{article}
\usepackage{codefuse_tech_report}
\usepackage[colorlinks = true,
            linkcolor = blue,
            urlcolor  = blue,
            citecolor = blue,
            anchorcolor = blue]{hyperref}   
\usepackage{microtype}
\usepackage{xurl}
\usepackage{booktabs}
\usepackage{enumitem}
\usepackage{multicol}
\usepackage{CJKutf8}
\usepackage{siunitx}
\usepackage{floatflt}
\usepackage{graphicx}
\usepackage{wrapfig}
\usepackage{authblk}
\usepackage{lipsum}
\usepackage{algorithm}
\usepackage{algorithmicx}
\usepackage{algpseudocode}
\usepackage{subfigure}
\usepackage{multirow}
\usepackage{pifont}  

\algnewcommand{\LeftComment}[1]{\Statex \(\triangleright\) #1}

\usepackage{array}
\usepackage{amsmath}
\usepackage{amssymb}
\usepackage{mathtools}
\usepackage{amsthm}

\usepackage[capitalize,noabbrev]{cleveref}
\usepackage{adjustbox} 

\theoremstyle{plain}

\theoremstyle{definition}

\theoremstyle{remark}

\colmfinalcopy

\usepackage[utf8]{inputenc}
\usepackage[T1]{fontenc}
\usepackage{caption} 
\usepackage{adjustbox} 
\usepackage{fontawesome5}

\usepackage{color}
\usepackage[table]{xcolor}
\usepackage{soul} 

\definecolor{tred}{RGB}{251, 130, 132}
\definecolor{torange}{RGB}{247, 162, 116}
\definecolor{tyellow}{RGB}{251, 218, 140}
\definecolor{tgreen}{RGB}{127, 204, 181}
\definecolor{tblue}{RGB}{89, 177, 215}
\definecolor{insightblue}{RGB}{162, 210, 255}
\definecolor{questionred}{RGB}{255, 175, 204}

\title{C2LLM Technical Report: A New Frontier in Code Retrieval via Adaptive Cross-Attention Pooling}

\author{%
Jin Qin\thanks{Equal Contribution.}$^{\phantom{*},1}$
~~Zihan Liao$^{*,1}$
~~Ziyin Zhang$^{*,1,2}$
\\

\vspace{-6pt}
\bf
~~Hang Yu\thanks{Correspondence to: Hang Yu \textless hyu.hugo@antgroup.com\textgreater, Peng Di \textless dipeng.dp@antgroup.com\textgreater, Rui Wang \textless wangrui12@sjtu.edu.cn \textgreater.}$^{\phantom{\dagger},1}$ 
~~Peng Di$^{\dagger,1}$
Rui Wang$^{\dagger,2}$

\vspace{10pt}
$^1$Ant Group\ \ \ $^2$Shanghai Jiao Tong University\\

\vspace{10pt}
{\footnotesize
\texttt{\
$^1$\{qj431428,liaozihan.lzh,hyu.hugo,dipeng.dp\}@antgroup.com
$^2$\{daenerystargaryen,wangrui12\}@sjtu.edu.cn}}\\

\vspace{10pt}
\hspace{-10pt}\faGithub ~\url{https://github.com/codefuse-ai/CodeFuse-Embeddings}\\
\hspace{-10pt}~~~~~~~~\includegraphics[width=1em,height=1em]{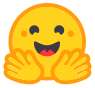} ~\href{https://huggingface.co/collections/codefuse-ai/codefuse-embeddings-68d4b32da791bbba993f8d14}{https://huggingface.co/collections/codefuse-ai/codefuse-embeddings}\\
}

\colmfinalcopy 

\begin{document}

\maketitle

\begin{figure}[h!]
    \centering
    \includegraphics[width=0.8\linewidth]{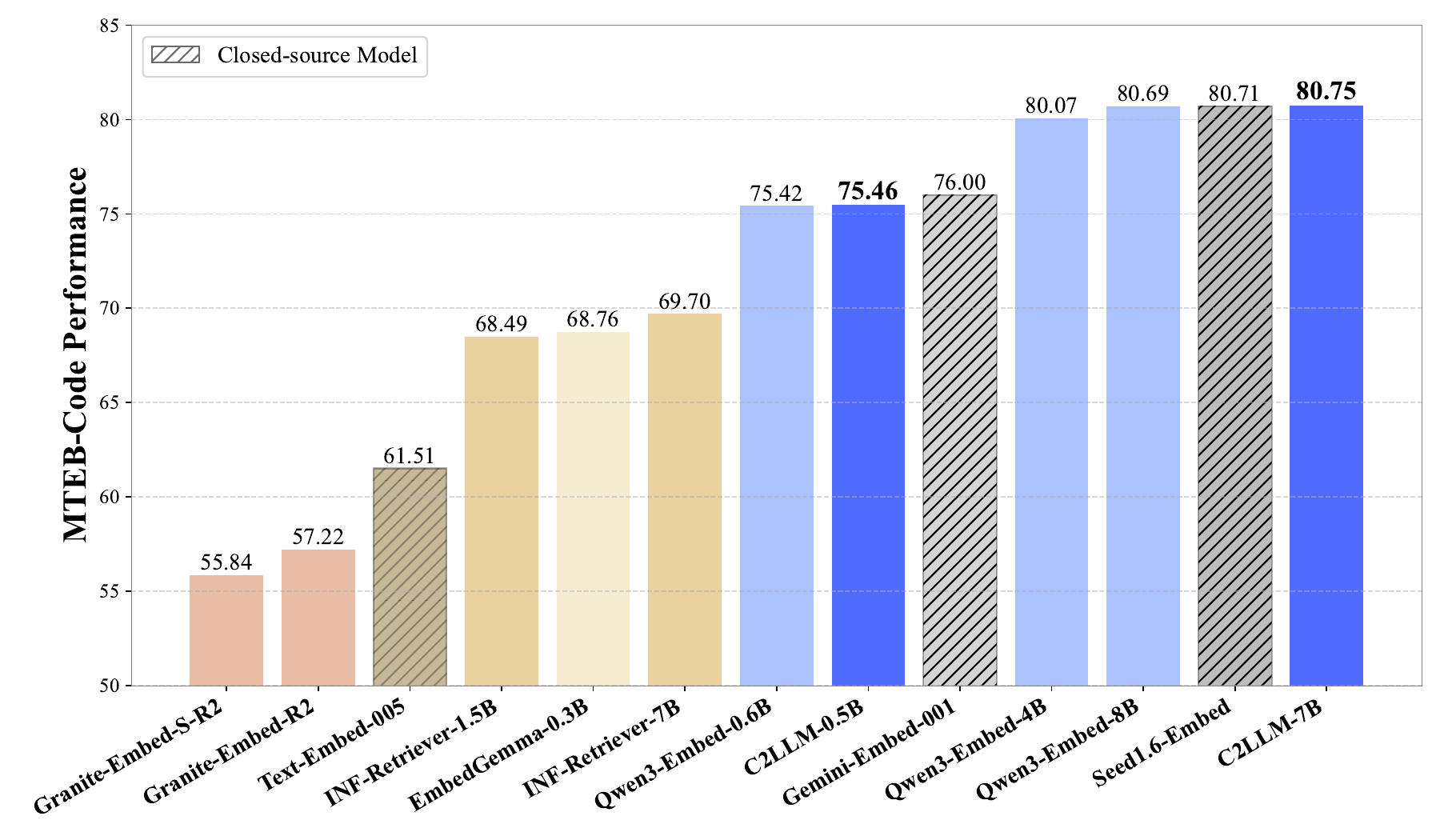}
    \caption{MTEB-Code leaderboard. C2LLM-7B ranks 1st among all models, surpasssing the best closed-source models, while C2LLM-0.5B ranks 1st among models with less than 1B parameters, and 6th overall.}
    \label{fig:performance}
\end{figure}

\begin{abstract}
We present C2LLM - Contrastive Code Large Language Models, a family of code embedding models in both 0.5B and 7B sizes. Building upon Qwen-2.5-Coder backbones, C2LLM adopts a Pooling by Multihead Attention (PMA) module for generating sequence embedding from token embeddings, effectively 1) utilizing the LLM's causal representations acquired during pretraining, while also 2) being able to aggregate information from all tokens in the sequence, breaking the information bottleneck in EOS-based sequence embeddings, and 3) supporting flexible adaptation of embedding dimension, serving as an alternative to MRL. Trained on three million publicly available data, C2LLM models set new records on MTEB-Code among models of similar sizes, with C2LLM-7B ranking 1st on the overall leaderboard.
\end{abstract}

\input{sec1-introduction}
\input{sec2-related-workd}
\input{sec3-model}

\input{sec4-experiments}
\input{sec5-conclusion}

\bibliographystyle{colm2024_conference}
\bibliography{custom}

\appendix
\clearpage

\end{document}

%% file: sec1-introduction.tex
\section{Introduction}

Large language models (LLMs) pretrained on source code and natural language have rapidly advanced a wide spectrum of software engineering applications, including code generation, automated issue resolution, and, notably, code retrieval~\citep{2024CodeLLMSurvey}. In the retrieval setting, a user supplies a natural-language query (e.g., ``open a jsonl file in Python and read all lines''), and the system must return the most relevant snippet among millions or even billions of candidates stored in public or private codebases. Code retrieval is not only essential for interactive developer search engines but also forms a pivotal step in the workflow of emerging code agents - autonomous systems that iteratively plan, search, and edit code to accomplish complex programming tasks~\citep{2024SWE-Agent,2025TraeAgent,2025CGM,2025OpenHands}.

At the core of code retrieval systems lie code embedding models. Despite the recent surge of general-purpose text embedding models~\citep{2025Qwen3-Embedding,2025NV-Embed,2025LGAI-Embedding-Preview,2024BGE-M3,2025F2LLM}, directly transferring them to code embedding remains sub-optimal, as \textbf{popular pooling strategies are ill-suited to code}. State-of-the-art embedding models either adopt mean pooling over the outputs of an LLM~\citep{2025NV-Embed,2025Gemini-Embedding} or take the end-of-sequence (EOS) token representation as sequence embeddings~\citep{2025LGAI-Embedding-Preview,2025F2LLM}. However, mean pooling is often paired with bidirectional attention, departing from the causal pretraining recipe of leading code LLMs~(e.g. Qwen2.5-Coder, \citealp{2024Qwen2.5-Coder}) and therefore fails to unlock their full potential~\citep{2025BGE-ICL}. Conversely, taking the EOS token embedding collapses all syntactic and semantic structure into one position, creating an information bottleneck that is especially harmful in the code domain, where input code files could easily contain thousands of tokens.

To address this challenge, we introduce Contrastive Code Large Language Models (C2LLM), a new code embedding model family optimized for code retrieval. \textbf{C2LLM preserves the causal attention of its backbone LLM but sidesteps the dilemma between mean pooling and EOS representation by inserting a lightweight Pooling by Multihead Attention (PMA) module}~\citep{2019SetTransformer}, which has been shown by \citet{D2LLM} to outperform both mean pooling and EOS representation. A single learnable query attends to all token representations produced by the LLM, simultaneously 1) aggregating sequence information into a single vector, and 2) providing support for dimensionality adaptation, making it ideal for real-world large-scale vector databases.

Trained on 3 million publicly available data, our 7B model achieves an average performance of 80.75 on MTEB-Code benchmark, ranking 1st among all models on the leaderboard. Our smaller model, with 0.5B parameters, scores 75.46 and pushes the frontier of models around 1B size, surpassing similar-sized competitors including Qwen3-Embedding-0.6B, EmbeddingGemma, and INF-Retriever. Our models are publicly available.








%% file: sec2-related-workd.tex
\section{Related Work}

In contrast to the abundance of text embedding models~\citep{2025NV-Embed,2025Qwen3-Embedding,2025F2LLM}, code-focused embedding research has received less attention in recent years. Most code embedding models adopt a BERT-based architecture, including CodeBERT~\citep{2020CodeBERT}, GraphCodeBERT~\citep{2021GraphCodeBERT}, CodeSage~\citep{2024CodeSage}, and CodeT5+~\citep{2023CodeT5p}, which fail to utilize the power of Code LLMs pretrained on trillions of tokens. BGE-Code~\citep{2025BGE-Code} and CodeXEmbed~\citep{2024CodeXEmbed} represent two notable exceptions, which are based on Qwen2.5-Coder and Mistral. However, none of these models are present on the MTEB-Code leaderboard, which is dominated by general-purpose text embedding models such as Qwen3-Embedding~\citep{2025Qwen3-Embedding}, INF-Retriever~\citep{inf-retriever}, and  EmbeddingGemma~\citep{2025EmbeddingGemma}.

%% file: sec3-model.tex
\section{Model Architecture: Introducing Pooling by Multihead Attention into Embedding Models}

\begin{figure}[th]
    \centering
    \includegraphics[width=1\linewidth]{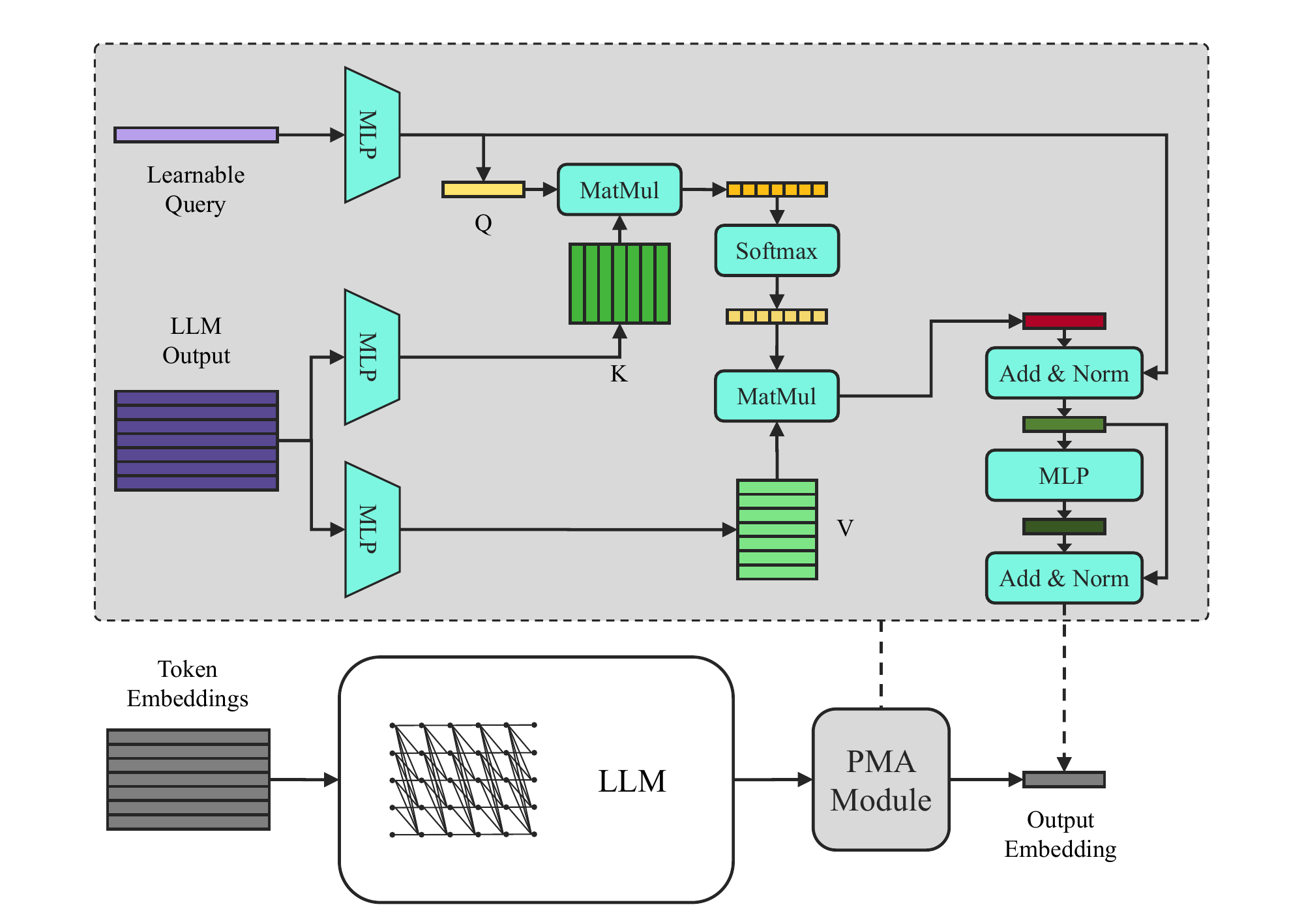}
    \caption{C2LLM Model architecture, comprising an LLM followed by a PMA (Pooling by Multihead Attention) module. PMA is a single layer of cross attention with one learnable query and takes the LLM's last hidden states as KV, serving both to pool over the input sequence and to provide support for flexible embedding dimension. Multi-head mechanism is omitted in the illustration.}
    \label{fig:model}
\end{figure}


Two of the most popular methods for obtaining an embedding from a token sequence are mean pooling~\citep{2025Gemini-Embedding,2025KaLM-Embedding-V2,2025QZhou-Embedding} and taking the EOS token embedding~\citep{2025Qwen3-Embedding,2025LGAI-Embedding-Preview}. However, mean pooling is often paired with bidirectional attention, deviating from state-of-the-art LLMs' pretraining design and thus being unable to fully exploit their potential~\citep{2025BGE-ICL}, while EOS representation condenses information from the entire sequence into a single token, creating an information bottleneck. To circumvent this dilemma, NV-Embed~\citep{2025NV-Embed} introduced a latent attention layer on top of the LLM, using the LLM's hidden states as query and a latent array of 512 vectors as key/value. This design, however, does not change the number of tokens and still requires mean pooling on the output.

In C2LLM, we propose yet another solution by introducing Pooling by Multihead Attention~(PMA, \citealp{2019SetTransformer,D2LLM}). As illustrated in Figure~\ref{fig:model}, PMA consists of a cross-attention layer with a single learnable query vector and takes the LLM's last hidden states as key/value, effectively aggregating information from the token sequence into a single embedding vector. Apart from pooling over the sequence dimension, PMA can also reduce the embedding dimension at the same time, providing an alternative to MRL~\citep{2022MRL}.

Formally, given the LLM's hidden states for $l$ input tokens $H\in\mathbb R^{l\times d_{\text{LLM}}}$ and the learnable query vector $q\in\mathbb R^{1\times d_q}$, we first project them into lower dimensions:
\begin{equation}
    Q^{1\times d} = qW_q,
\end{equation}
\begin{equation}
    K^{l\times d} = HW_k,
\end{equation}
\begin{equation}
    V^{l\times d} = HW_v,
\end{equation}

where $W_q\in\mathbb R^{d_q\times d}$, $W_k, W_v\in\mathbb R^{d_{\text{LLM}}\times d}$, and $d$ is the output embedding dimension. Cross attention is then computed in this lower dimension with residual connections and layer normalization (LN):
\begin{equation}
    O^{1\times d} = \text{softmax}(QK^T)V,
\end{equation}
\begin{equation}
    \tilde O^{1\times d} = \text{LN}(O + Q),
\end{equation}
\begin{equation}
    E^{1\times d} = \text{LN}(\text{ReLU}(\tilde OW_o) + \tilde O).
\end{equation}

$E$ is then taken as the embedding for the input sequence.

\paragraph{Takeaway} The integration of the PMA module into embedding models offers three primary advantages. First, unlike mean pooling and EOS representation, the cross-attention mechanism allows the model to learn which tokens (e.g., function signatures or key algorithmic logic) are most salient for the final representation. Second, it maintains both the foundational causal architecture and efficiency of the LLM backbone, as the PMA overhead is negligible compared to the billions of parameters in the LLM. Finally, by decoupling the LLM's hidden dimension ($d_{\text{LLM}}$) from the final embedding dimension ($d$), PMA can produce compact embeddings suitable for vector databases without requiring the computationally expensive MRL training objective.








%% file: sec4-experiments.tex
\section{Experiments}

\subsection{Training Settings}

\textbf{Model Configurations}
We develop the C2LLM series by fine-tuning two state-of-the-art base models: Qwen2.5-Coder-0.5B-Instruct and Qwen2.5-Coder-7B-Instruct~\citep{2024Qwen2.5-Coder}. The training data includes CodeSearchNet~(including code-to-code, code-to-text, and text-to-code retrieval, \citealp{2019CodeSearchNet,2025CoIR,2021CodeXGLUE}), APPS~\citep{2021APPS}, single-turn and multi-turn CodeFeedback~\citep{2024OpenCodeInterpreter}, CodeEditSearch~\citep{2024OctoPact}, CosQA~\citep{2021CosQA}, StackOverflowQA~\citep{2025CoIR}, SyntheticText2SQL~\citep{2024synthetic-text-to-sql}, and CodeTransOcean~\citep{2023CodeTransOcean}, totaling $3$ million samples. For the model configurations, we employ PMA with $32$ heads to aggregate token-level features into a single sequence representation. The fine-tuning process is made efficient through the use of LoRA~\citep{2022LoRA}, configured with a rank ($r$) of $64$ and an alpha ($\alpha$) of $32$. To optimize computational throughput and memory usage, we utilize Flash Attention 2~\citep{2024FlashAttention2} across all training stages.

\textbf{Training Strategy}
The models are trained for $3$ epochs with a learning rate of $1 \times 10^{-4}$ and a maximum sequence length of $1024$ tokens using left-padding. Our optimization strategy centers on contrastive learning. For in-batch contrastive learning, we implement a global batch strategy to synchronize samples across all distributed processes, effectively expanding the pool of negative samples. For hard-negative contrastive learning, we incorporate $K=7$ hard negatives for each query. We apply a temperature scaling factor of $\tau = 0.05$ to both in-batch and hard-negative contrastive losses. To ensure the quality of the contrastive signals, we adopt a specialized batching strategy where data is grouped according to both the dataset source and the specific programming language before being partitioned into training batches. During the optimization process, a loss weight of $1$ is assigned to all objectives, with the sole exception of the CodeEditSearch dataset, which uses a custom weight to balance its contribution. Finally, the definitive C2LLM model is produced by performing a weighted merge of four checkpoints captured at different global steps, a technique designed to enhance the stability and generalization of the final embeddings.

\textbf{Prompt template}
The Prompt templates for each dataset are shown in Table~\ref{table:prompt}.

\input{tables/instruction}

\subsection{Results}
\input{tables/results}

We evaluate C2LLM on the $12$ retrieval tasks in MTEB-Code Benchmark~\citep{2023MTEB,2025MMTEB}\footnote{\url{https://huggingface.co/spaces/mteb/leaderboard}}. As shown in Table~\ref{tab:results}, C2LLM-$7$B achieves an average score of $80.75$, surpassing the previous state-of-the-art Seed$1.6$-Embedding and Qwen3-Embedding-$8$B. Notably, C2LLM-$7$B shows superior performance in complex reasoning tasks such as CodeFeedback ($94.32$ for multi-turn, $90.66$ for single turn), suggesting that the PMA module effectively captures the intent behind natural language queries directed at code.

Our smaller variant, C2LLM-$0.5$B, demonstrates remarkable efficiency. With only $0.5$B parameters, it achieves an average score of $75.46$, outperforming significantly larger models like INF-Retriever-$7$B ($69.70$). It also surpasses all other models with less than $1$B parameters, establishing a new state-of-the-art in the compute-efficient regime. The consistent performance of C2LLM across both scales validates the robustness of using cross-attention as a universal pooling strategy for code embeddings.

%% file: tables/instruction.tex
\begin{table}[t]
    \centering
    \adjustbox{width=\textwidth+0.1cm,center}{
    \rowcolors{2}{gray!10}{white}
    \begin{tabular}{lm{15cm}c}
\toprule
Task Name & Query Instruction &  Document Instruction \\
\midrule
CodeEditSearchRetrieval    &  Retrieve the diff code that relevant the following query:& Retrieved Answer: \\

CodeSearchNetRetrieval    &    Retrieve the code that solves the following query: & Retrieved Answer: \\

AppsRetrieval    &    Given a problem description from a programming contest, retrieve code examples that can assist in solving it.& Retrieved Answer: \\

CodeFeedbackMT    &   Given a multi-turn conversation history that includes both text and code, retrieve relevant multi-modal answers composed of text and code that address the ongoing discussion. & Retrieved Answer: \\

CodeFeedbackST    &     Given a single-turn question composed of text and code, retrieve suitable answers that also mix text and code to provide helpful feedback. & Retrieved Answer: \\

CodeSearchNetCCRetrieval    &     Given an initial code segment, retrieve the subsequent segment that continues the code. & Retrieved Answer: \\

CodeTransOceanContest    &     Given a Python code snippet, retrieve its semantically equivalent version written in C++. & Retrieved Answer: \\

CodeTransOceanDL    &     Given a code snippet, retrieve a semantically equivalent implementation of the same code. & Retrieved Answer: \\

COIRCodeSearchNetRetrieval    &     Given a code snippet, retrieve its corresponding document string that summarizes its functionality. & Retrieved Answer: \\

CosQA    &     Given a query from a web search, retrieve code that is helpful in addressing the query. & Retrieved Answer: \\

StackOverflowQA    &     Given a question combining text and code, retrieve relevant answers that also contain both text and code snippets and can address the question. & Retrieved Answer: \\

SyntheticText2SQL    &    Given a natural language question, retrieve SQL queries that serve as appropriate responses. & Retrieved Answer: \\
\bottomrule
    \end{tabular}
    }
    \caption{Instructions for training data.}
    \label{table:prompt}
\end{table}

%% file: tables/results.tex
\begin{table}[t]
    \centering
    \tiny
    \adjustbox{width=\textwidth+0.1cm,center}{
    \begin{tabular}{crc>{\centering}m{1.5cm}>{\centering}m{0.75cm}>{\centering}m{1.15cm}>{\centering}m{1.4cm}>{\centering}m{0.5cm}>{\centering}m{0.9cm}>{\centering}m{1.0cm}>{\centering}m{0.35cm}c}
    \toprule
        Model & Size & APPS & CodeSearchNet (CSN/CCR/CoIR) & CodeEdit & CodeFeedback (MT/ST) & CodeTransOcean (Contest/DL) & CosQA & Stack-OverflowQA & Synthetic-Text2SQL & Avg & Rank \\
    \midrule
        \rowcolor{gray!15}
        C2LLM & 7B & 86.71 & 91.07/97.90/89.79 & 81.49 & 94.32/90.66 & 92.51/34.13 & 39.76 & 94.85 & 75.75 & 80.75 & 1 \\
        Seed1.6-Embed & NA & 91.15 & 93.17/94.15/89.50 & 92.14 & 90.11/90.44 & 90.16/35.99 & 41.17 & 97.20 & 63.31 & 80.71 & 2 \\
        Qwen3-Embed & 8B & 91.07 & 92.66/96.35/89.51 & 76.97 & 93.70/89.93 & 93.73/32.81 & 38.04 & 94.75 & 78.75 & 80.69 & 3 \\
        Qwen3-Embed & 4B & 89.18 & 92.34/95.59/87.93 & 76.49 & 93.21/89.51 & 90.99/35.04 & 37.98 & 94.32 & 78.21 & 80.07 & 4 \\
        Gemini-Embed & NA & 93.75 & 91.33/84.69/81.06 & 81.61 & 56.28/85.33 & 89.53/31.47 & 50.24 & 96.71 & 69.96 & 76.00 & 5 \\
        \rowcolor{gray!15}
        C2LLM & 0.5B & 61.02 & 89.20/96.29/86.71 & 71.39 & 92.29/88.63 & 84.27/33.99 & 38.30 & 89.40 & 74.08 & 75.46 & 6 \\
        Qwen3-Embed & 0.6B & 75.34 & 91.01/91.72/84.69 & 64.42 & 90.82/86.39 & 86.05/31.36 & 36.48 & 89.99 & 76.74 & 75.42 & 7 \\
        INF-Retriever & 7B & 47.37 & 88.77/75.71/72.27 & 71.79 & 77.64/86.63 & 89.16/35.18 & 34.18 & 94.22 & 63.51 & 69.70 & 8 \\
        EmbedGemma & 0.3B & 84.39 & 90.15/73.71/75.54 & 62.10 & 51.42/80.26 & 85.51/33.52 & 43.60 & 86.47 & 58.42 & 68.76 & 9 \\
        INF-Retriever & 1.5B & 38.90 & 90.87/75.50/78.63 & 67.17 & 77.47/84.51 & 85.01/33.84 & 33.11 & 91.32 & 65.59 & 68.49 & 10 \\
    \bottomrule
    \end{tabular}
    }
    \caption{Top $10$ models on the MTEB-Code leaderboard as of the submission date (2025-12-25). ``NA'' in the model size column indicates closed-source model whose size is not available.}
    \label{tab:results}
\end{table}

%% file: sec5-conclusion.tex
\section{Conclusion}


We introduce C2LLM, a family of code embedding models that achieves state-of-the-art performance by combining the strengths of causal LLM pretraining with a flexible Pooling by Multihead Attention (PMA) module. Our results demonstrate that bypassing the historical dilemma between EOS and mean-pooling strategies allows for better information aggregation in representing code sequences, setting new records on the MTEB-Code benchmark with our 7B model.

C2LLM represents the fourth entry in the CodeFuse Embedding model family, following D2LLM~\citep{D2LLM}, E2LLM~\citep{2025E2LLM}, and F2LLM~\citep{2025F2LLM}. We are dedicated to promoting open research in LLM-based embedding models, and plan to expand the series into massively multilingual and multi-domain scenarios in the near future.
